\documentclass{article} %
\usepackage{colm2024_conference}

\usepackage{microtype}
\usepackage{hyperref}
\usepackage{booktabs}
\usepackage{multirow}
\usepackage{cleveref}
\usepackage{xspace}
\usepackage{graphicx}
\usepackage{makecell}
\usepackage{url}
\usepackage{caption}

\def\modelname{MaLA-500\xspace}
\def\sib{SIB200\xspace}
\def\taxi{Taxi1500\xspace}
\def\nll{$NLL$\xspace}
\def\langnum{534\xspace}

\title{\modelname: Massive Language Adaptation of Large Language Models}

\author{Peiqin Lin$^*$$^{1,2}$, Shaoxiong Ji$^*$$^{3}$, Jörg Tiedemann$^{3}$, André F. T. Martins$^{4,5,6}$, Hinrich Schütze$^{1,2}$ \\
        $^1$Center for Information and Language Processing, LMU Munich \\
        $^2$Munich Center for Machine Learning \quad
        $^3$University of Helsinki \\
        $^4$Instituto Superior Técnico (Lisbon ELLIS Unit) \\
        $^5$Instituto de Telecomunicações \quad
        $^6$Unbabel \\
        \texttt{linpq@cis.lmu.de, shaoxiong.ji@helsinki.fi}
}

\colmfinalcopy %
\begin{document}

\def\thefootnote{*}\footnotetext{Equal contribution.}\def\thefootnote{\arabic{footnote}}

\maketitle

\begin{abstract}
Large language models (LLMs) have advanced the state of the art in natural language processing.
However, their predominant design for English or a limited set of languages creates a substantial gap in their effectiveness for low-resource languages.
To bridge this gap, we introduce \modelname, a novel large language model designed to cover an extensive range of 534 languages. To train \modelname, we employ vocabulary extension and continued pretraining on LLaMA 2 with Glot500-c.
Our intrinsic evaluation demonstrates that \modelname is better at predicting the given texts of low-resource languages than existing multilingual LLMs.
Moreover, the extrinsic evaluation of in-context learning shows that \modelname outperforms previous LLMs on \sib and \taxi by a significant margin, i.e., 11.68\% and 4.82\% marco-average accuracy across languages.
We release \modelname at \url{https://huggingface.co/MaLA-LM}.
\end{abstract}

\section{Introduction}

Large Language Models (LLMs), e.g., LLaMA \citep{touvron2023llama,DBLP:journals/corr/abs-2307-09288}, Mistral \citep{DBLP:journals/corr/abs-2310-06825,jiang2024mixtral}, and ChatGPT,\footnote{\url{https://openai.com/blog/chatgpt}} have shown remarkable performance in natural language understanding and generation.
Follow-up studies \citep{DBLP:journals/corr/abs-2302-04023,DBLP:journals/corr/abs-2304-05613,DBLP:journals/corr/abs-2303-12528,DBLP:journals/corr/abs-2311-07463} observe that these English-centric LLMs, such as LLaMA with mainly English as the training data, are capable of handling some high-resource non-English languages, benefiting from the inclusion of non-English language data during pretraining. However, their applicability to low-resource languages is still limited due to data scarcity. 

Previous studies have released pretrained multilingual models with mostly encoder-only transformer architectures, e.g., multilingual BERT~\citep{devlin2019bert} and XLM-R~\citep{conneau2020unsupervised}, for around 100 languages. 
The paradigm shift from encoder-only to decoder-only achieves scalability for large language models with billions of model parameters, leading to the development of open multilingual models. 
Recently, several generative multilingual LLMs, such as XGLM \citep{DBLP:journals/corr/abs-2112-10668}, mGPT \citep{DBLP:journals/corr/abs-2204-07580}, and BLOOM~\citep{scao2022bloom}, have emerged.
Notably, the current language coverage for these generative LLMs is limited to up to 60 languages, highlighting the remaining need for further work on massively multilingual LLMs for many natural languages.

\citet{imanigooghari-etal-2023-glot500} have achieved a significant milestone in the realm of massive language adaptation by extending the language coverage of a small-scale multilingual language model, XLM-R~\citep{conneau2020unsupervised} - an auto-encoding model with 278M parameters, from 100 languages to an impressive number of \langnum languages, and introducing an extended model, Glot500-m with 395M parameters. 
\citet{imanigooghari-etal-2023-glot500} introduce the Glot500-c corpora spanning \langnum languages from 47 language families, and then apply vocabulary extension and continued pretraining to create Glot500-m. The introduction of Glot500-c mitigates the challenge of data scarcity for low-resource languages. Moreover, the adaptation method is more favorable than training from scratch, as it requires fewer computational resources and emits a smaller carbon footprint. This success serves as a strong motivation for our exploration into the massive language adaptation of LLMs.

This work aims to extend the capabilities of LLMs to encompass a wider range of languages.
Existing works like~\citet{imanigooghari-etal-2023-glot500} on language adaptation of pretrained models provide extended coverage across a wide linguistic spectrum but are limited to relatively small model sizes - mostly at the hundred million scales, while other works like~\citet{DBLP:journals/corr/abs-2212-09535} extended generative LLMs but are limited to a small number of languages. 
Our study pushes the boundaries by exploring language adaptation techniques for LLMs with model parameters scaling up to 10 billion for \langnum languages.
Our investigation delves into generative LLMs with a substantial increase in model parameters and their in-context learning capabilities in diverse languages, especially low-resource languages.
This augmentation enables us to enhance contextual and linguistic relevance across a diverse range of languages.

We address the challenges of adapting LLMs to low-resource languages, such as data sparsity, domain-specific vocabulary, and linguistic diversity. 
Specifically, we study continued pretraining of open LLM, i.e., LLaMA 2~\citep{DBLP:journals/corr/abs-2307-09288}, vocabulary extension, and adaptation techniques, i.e., LoRA low-rank reparameterization~\citep{hu2022lora}.
We deploy distributed training and release \modelname that covers more than 500 languages in various domains. 
We evaluate \modelname using intrinsic measures on held-out Glot500-c test set and parallel data and extrinsic metrics on downstream benchmarks: \sib and \taxi.
The results show that \modelname outperforms existing open LLMs of close or slightly larger model size.
This work broadens the accessibility of LLMs, making them valuable for a more diverse set of language-specific use cases, especially for low-resource ones, and addressing the equality issue by removing language barriers for speakers of many languages, especially those underrepresented languages covered by existing LLMs.

\section{Massive Language Adaptation}

The principle of massive language adaptation of large language models accommodates the utilization of a massively multilingual corpus (\Cref{sec:data}), the strong base LLM (\Cref{sec:model}), and the technique for effective language adaptation: vocabulary extension (\Cref{sec:vocabulary_extension}) and continued pretraining (\Cref{sec:continual_training}).

\subsection{Data}
\label{sec:data}

We use Glot500-c~\citep{imanigooghari-etal-2023-glot500} covering \langnum languages\footnote{We define languages using the ISO 639-3 code combined with the corresponding written script. For example, ``\texttt{eng\_Latn}'' represents English written in the Latin script.} as the training data of \modelname.  See \S\ref{sec:appendix_languages} for the list of languages with their data amounts. 
The original number of sentences ranges from 10 thousand to 63 million. 
Note that Glot500-c does not put full effort into collecting data for high-resource languages but focuses on low-resource languages.
We sample languages from the imbalanced dataset according to a multinomial distribution, with $\alpha=0.3$ for vocabulary extension and continued pretraining.
We use different scales for sampling data to be used in model training and vocabulary construction. 
After sampling, the number of sentences for training ranges from 600 thousand to 8 million per language, leading to 1 billion sentences in total.
The number of sentences for vocabulary construction ranges from 30 thousand to 400 thousand, making a total of 50 million sentences. 

\subsection{Model}
\label{sec:model}

We choose LLaMA 2~\citep{DBLP:journals/corr/abs-2307-09288} to start continual training.
LLaMA series models~\citep{touvron2023llama}, with model weights released publicly, have gained popularity in the research community.
Despite being English-centric compared to their multilingual counterparts, they have shown remarkable capacity for multiple languages~\citep{DBLP:journals/corr/abs-2311-07463}. 
We choose the latest LLaMA 2, trained on 2 trillion tokens, as our base model to benefit from its outstanding language capacity. 
Our study chooses the 7B model with 32 transformer layers, and leaves the extension of LLMs with larger sizes as a future work.

\subsection{Vocabulary Extension}
\label{sec:vocabulary_extension}

The original LLaMA 2's 32,000 tokenizer covers English and a small fraction of other European languages using Latin or Cyrillic scripts. To enhance its capability and encoding efficiency for a broader range of languages, we extend the vocabulary with Glot500-c. Specifically, we initially train a multilingual tokenizer with SentencePiece \citep{DBLP:conf/emnlp/KudoR18} on the sampled Glot500-c with a vocabulary of 250,000. Subsequently, we merge the trained tokenizer with the original LLaMA 2 tokenizer by taking the union of their vocabularies. As a result, we obtain the \modelname's tokenizer with a vocabulary size of 260,164. 
After vocabulary extension and resizing the embedding layer, the model size becomes 8.6B. 

We measure the impact of vocabulary extension on the development set of Glot500-c by analyzing the reduction in segmentation length for each language. The results indicate that the effect of vocabulary extension varies, ranging from 8\% (English, \texttt{eng\_Latn}) to 88\% (Oriya, \texttt{ori\_Orya}). Unsurprisingly, vocabulary extension has a larger effect on languages written in non-Latin scripts than on those in the Latin script. However, for some low-resource languages written in the Latin script, e.g., Kabiyè (\texttt{kbp\_Latn}) and Vietnamese (\texttt{vie\_Latn}), the segmentation length is shortened by around 50\%.

\subsection{Continued Pretraining}
\label{sec:continual_training}
We employ continued pretraining for language adaptation with low-rank adaptation~\citep[LoRA,][]{hu2022lora} to enable parameter-efficient training, given the limitation of our computing resources. 
LoRA injects trainable rank decomposition matrices, which approximate the large weight matrices with a lower rank, to the pretrained model weights. 
It reduces the computational complexity and thus saves the training cost while retaining high model quality~\citep{hu2022lora}.
We continually train the casual language model to update the rank-decomposition matrices, embedding layer, and language modeling head while freezing the transformer weights of pretrained models, allowing the continually trained language model to learn from new data in new languages without completely losing its previous language capacity.
Continual training of large language models requires substantial computational resources. 
We adopt efficient distributed training setups on supercomputers to make the training process feasible.

\subsection{Training}
\label{sec:training}

\paragraph{Hardware and Software}
We train our model on the computing cluster with the theoretical peak performance of 2 petaflops on GPU nodes.
We deploy distributed training on 24 Nvidia Ampere A100 GPUs. 
As for software, we utilize the Huggingface Transformers~\citep{wolf2020transformers}, PEFT (Parameter-Efficient Fine-Tuning),\footnote{\url{https://huggingface.co/docs/peft/index}} 
and DeepSpeed~\citep{rasley2020deepspeed}. 
We use the ZeRO redundancy optimizer~\citep{rajbhandari2020zero} and maximize the batch size that fits the memory of each GPU. 
We employ mixed-precision training using the bfloat16 format.

\paragraph{Hyperparameters}

The learning rate is set at 3e-4. 
A weight decay of 0.01 is applied to penalize large weights and mitigate overfitting. 
The trainable LoRA module targets the query and value matrices.
The language model head is not decomposed by a LoRA module but is trained in a full-parameter manner. 
In our setting, the final model has 10B parameters in total, in which 2B parameters are trainable.
The LoRA module is incorporated with a rank of 8, an alpha value of 32, and a dropout rate of 0.05, contributing to the model's adaptability and regularization during training.
The context window is 4k.
We maximize the batch size to fit the memory, making a global batch size of 384.
The model undergoes three training epochs.
Checkpoints are saved every 500 steps, and we employ early stopping to select the checkpoint that exhibits the most favorable average performance on downstream tasks.

\paragraph{Environmental Impacts}

We train our model on a carbon-neutral data center, with all electricity generated with renewable hydropower, and the waste heat is utilized in district heating to further reduce CO2 footprint.\footnote{\url{https://www.csc.fi/sustainable-development}}

\section{Evaluation}

\subsection{Benchmarks and Setup}
\label{sec:setup}

We consider both intrinsic and extrinsic measures for evaluation. Evaluation dataset statistics are shown in Table~\ref{tab:tasks}.

\begin{table}[ht]
    \centering
    \resizebox {\columnwidth} {!} {
    \begin{tabular}{lccccc}
    \toprule
      & Datasets & Metric & $\|$Data$\|$ & $\|$Lang$\|$ & Domain \\
    \midrule
    \multirow{2}{*}{Intrinsic} & Glot500-c test \citep{imanigooghari-etal-2023-glot500} & \nll & 1000 & 534 & Misc \\
     & PBC \citep{DBLP:conf/lrec/MayerC14} & \nll & 500 & 370 & Bible \\
    \midrule
     \multirow{2}{*}{Extrinsic}&\sib \citep{DBLP:journals/corr/abs-2309-07445} & ACC & 204 & 177 & Misc \\
     &Taxi1500 \citep{ma2023taxi1500} & ACC & 111 & 351 & Bible \\
    \bottomrule
    \end{tabular}
    }
    \caption{Evaluation dataset
    statistics.
    $\|$Data$\|$: test set size per language.
    $\|$Lang$\|$: number of evaluated languages. \nll: negative log-likelihood.
    ACC: Accuracy.}
    \label{tab:tasks}
\end{table}

For intrinsic evaluation, perplexity is not comparable across models and languages due to different text segmentations. Inspired by \citet{DBLP:journals/tacl/XueBCANKRR22,DBLP:journals/corr/abs-2305-07185}, we instead measure the negative log-likelihood (\nll) of the text using the given LLMs. %

We concatenate the dataset as the input text and adopt the sliding-window strategy.\footnote{\url{https://huggingface.co/docs/transformers/en/perplexity}}
The evaluation of different LLMs uses the same data with the concatenation of sentences per language, 
thus making \nll model-comparable.
In addition, we consider language-comparable \nll by measuring \nll on parallel data, in which every sample in different languages contains the same semantic information.
We report the model-comparable \nll of Glot500-c test set covering all \langnum considered languages (\S\ref{sec:model_cmp}), and language-comparable \nll on Parallel Bible Corpus ~\citep[PBC,][]{DBLP:conf/lrec/MayerC14}, covering 370 languages (\S\ref{sec:lang_cmp}).

For extrinsic evaluation, we evaluate the few-shot learning capability of \modelname and compare it with other LLMs on \sib \citep{DBLP:journals/corr/abs-2309-07445} and \taxi \citep{ma2023taxi1500}.

\sib is a topic classification dataset. The classification task involves seven classes, namely science/technology, travel, politics, sports, health, entertainment, and geography. Our evaluation spans a diverse set of 177 languages, obtained by intersecting the language sets of \sib and Glot500-c. 
Note that the flores200-based \sib evaluation set is included in the training data since Glot500-c includes flores200, but the classification labels are not provided.

\taxi is another text classification dataset spanning 351 languages. It involves six classes, namely, Recommendation, Faith, Description, Sin, Grace, and Violence. 
Our evaluation efforts aim to cover as many languages as possible. 
However, the evaluation of massively multilingual language models is a challenging task. Due to the lack of real-world multilingual evaluation benchmarks, we use this benchmark that contains religious content. 

For in-context learning evaluation, the evaluated LLM receives a structured prompt, which is the concatenation of few-shot examples and the sample intended for prediction. The format for both a few-shot example and the sample to predict is defined as follows:

Template for \sib:
\begin{quote}
    The topic of the news [sent] is [label]
\end{quote}
Template for \taxi:
\begin{quote}
    The topic of the verse [sent] is [label]
\end{quote}
where [sent] is the sentence for classification, and [label] is the ground truth. [label] is included when the sample serves as a few-shot example but is omitted when predicting the sample. The constructed prompt is then used as input to the LLM. Subsequently, the evaluated LLM is prompted to estimate the probability of the label over the label set based on the provided prompt.

For \sib, few-shots examples are randomly sampled from the in-language training sets. Since randomly selecting few-shot examples for in-context learning yields random results for both \modelname and previous LLMs on \taxi, we consider the retriever-based in-context learning \citep{DBLP:conf/acl-deelio/LiuSZDCC22}. Specifically, we use average word embeddings in layer 8 of the Glot500 \citep{imanigooghari-etal-2023-glot500} for retrieving semantic-similar samples as suggested in previous work \citep{DBLP:conf/emnlp/SabetDYS20} for all the compared models. The evaluation process is implemented using the lm-evaluation-harness,\footnote{\url{https://github.com/EleutherAI/lm-evaluation-harness}} and we use accuracy (ACC) to measure the performance of classification.

\subsection{Comparison across LLMs}
\label{sec:model_cmp}
We compare \modelname with LLaMA 2-7B, mGPT-13B, BLOOM-7B1, and XGLM-7.5B on Glot500-c test set, \sib, \taxi by computing the averaged performance across languages, and the result are given in \Cref{tab:comparison_across_LLMs}.
Among the evaluated LLMs, LLaMA 2-7B performs second-best, indicating that LLaMA 2-7B has a strong multilingual capacity and that it is reasonable to select it as the base model. \modelname outperforms all compared LLMs with a close or slightly larger model size across all the evaluated tasks. Notably, compared to LLaMA 2-7B, \modelname gains a lower \nll on the Glot500-c test set by 39.33, and has 14.94\% and 4.82\% improvements on \sib and \taxi, respectively. It highlights \modelname's substantial contribution to enhancing the multilingual capacity of LLMs.

\begin{table}[ht]
\begin{minipage}[b]{\linewidth}
    \centering
    \small
    \begin{tabular}{c|ccc}
    \toprule
    Model & Glot500-c test (\nll$\downarrow$) & \sib (ACC $\uparrow$) & \taxi (ACC $\uparrow$)\\
    \midrule
    LLaMA 2-7B & 190.58 & 42.08 & 44.07 \\
    mGPT-13B & 282.46 & 45.34 & 40.98 \\
    BLOOM-7B1 & 202.95 & 44.63 & 43.98 \\
    XGLM-7.5B & 205.07 & 34.36 & 43.24 \\
    \midrule
    \modelname & \textbf{151.25} & \textbf{57.02} & \textbf{48.89} \\
    \bottomrule
    \end{tabular}
    \caption{Averaged results across languages on Glot500-c test (measured by \nll), \sib, and \taxi (measured by accuracy (\%)) of different LLMs. mGPT has no model with around 7B parameters, so we choose a larger one with 13B parameters. $\downarrow$ indicates the lower, the better. $\uparrow$ indicates the higher, the better. The best results are \textbf{bold}.}
    \label{tab:comparison_across_LLMs}
\end{minipage}
\end{table}

\begin{figure}[ht]
    \centering
    \resizebox {0.6\columnwidth} {!} {
    \includegraphics[clip]{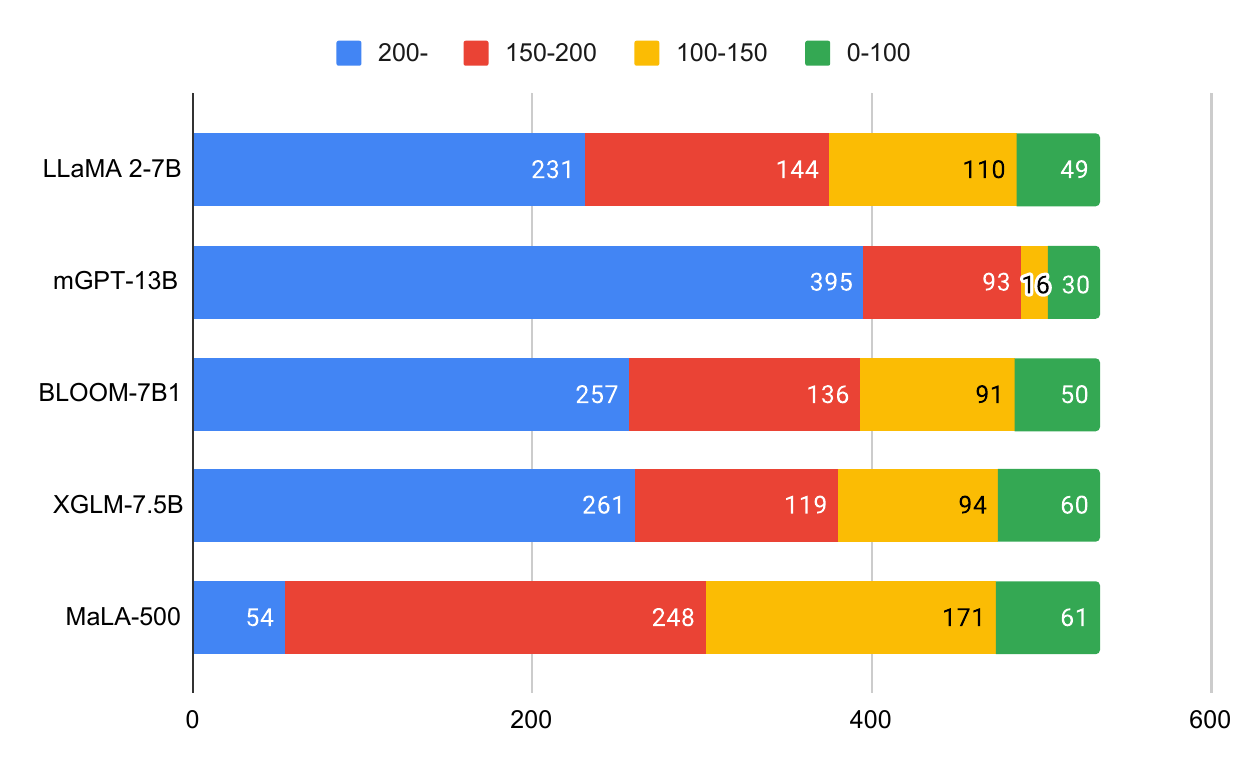}
    }
    \captionof{figure}{\nll (lower is better) on Glot500-c test with the scores grouped into four bins displayed in different colors. X-axis: the number of languages in performance ranges.}
    \label{fig:comparison_across_LLMs_nll}
\end{figure}

\begin{figure}[ht]
    \centering
    \resizebox {0.6\columnwidth} {!} {
    \includegraphics[clip]{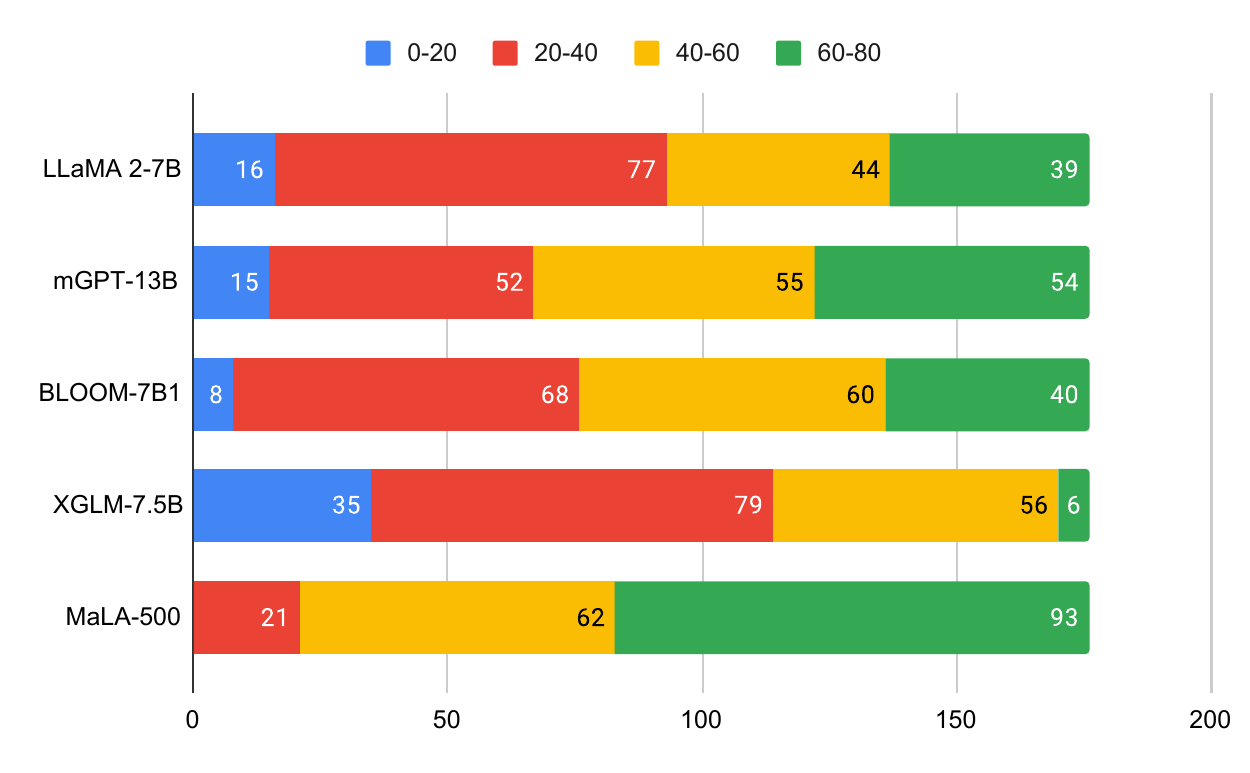}
    }
    \captionof{figure}{Accuracy (higher is better) on \sib with the scores grouped into four bins displayed in different colors. X-axis: the number of languages in performance ranges (\%).}
    \label{fig:comparison_across_LLMs_sib200}
\end{figure}

\begin{figure}[ht]
    \centering
    \resizebox {0.6\columnwidth} {!} {
    \includegraphics[clip]{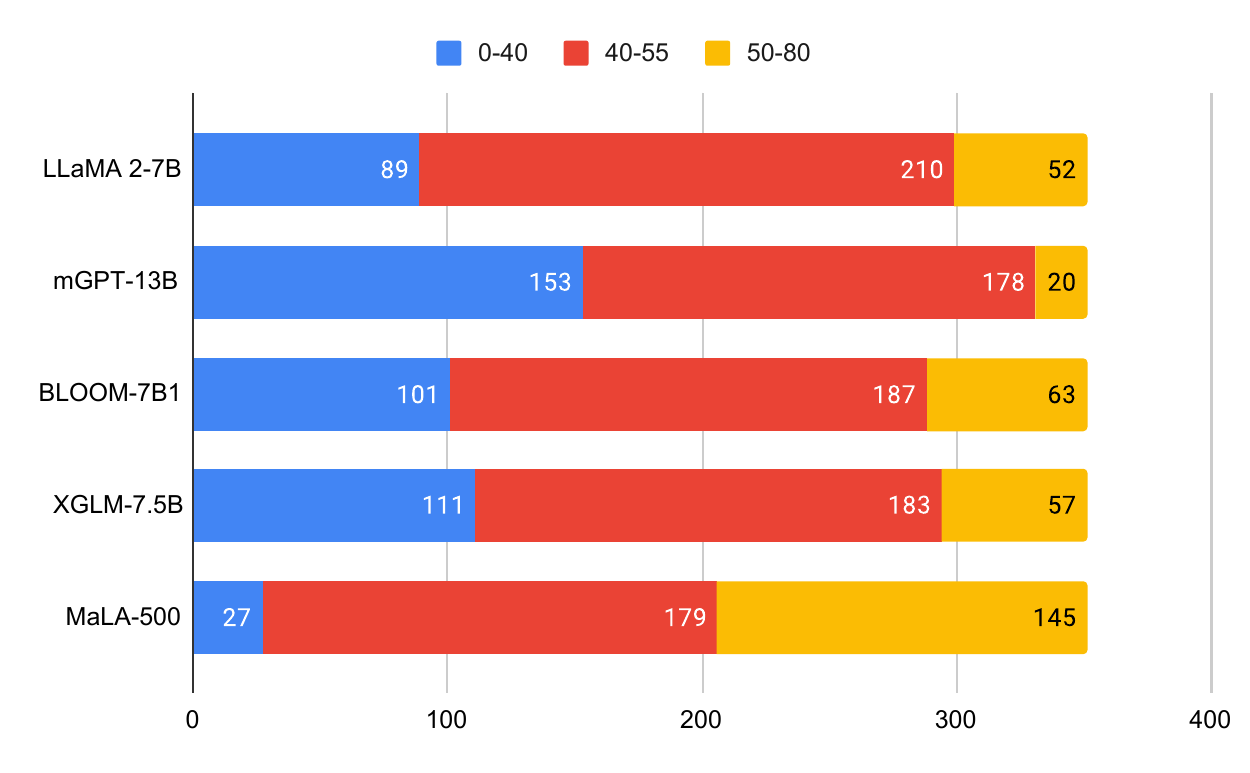}
    }
    \captionof{figure}{Accuracy (higher is better) on \taxi with the scores grouped into four bins displayed in different colors. X-axis: the number of languages in performance ranges (\%).}
    \label{fig:comparison_across_LLMs_taxi}
\end{figure}

\Cref{fig:comparison_across_LLMs_nll,fig:comparison_across_LLMs_sib200,fig:comparison_across_LLMs_taxi} provide detailed performance analysis across languages on Glot500-c test, \sib, and \taxi. 
In those figures, we group scores into different performance bins and display them in different colors. 
For Glot500-c test, \modelname has more languages achieving better \nll, i.e., 61 languages with \nll less than 100 and 171 languages with \nll between 100 and 150. Besides, \modelname has 54 (10\%) languages achieving \nll larger than 200, which may indicate the languages are not well covered by the measured LLM. Nevertheless, the number is much less than other LLMs. For example, the second-best LLM, LLaMA 2-7B, has 231 (43\%) languages achieving \nll larger than 200.
For both \sib and \taxi, \modelname surpasses previous LLMs 
in the sense that it obtains random results in fewer languages and achieves impressive performance in more languages than its counterparts.

\subsection{Comparison across Languages}
\label{sec:lang_cmp}

To check in detail how \modelname performs across languages, we check the performance across language families\footnote{We
assign languages to families based on Glottolog: \url{https://glottolog.org/glottolog/family}.} shown in Table~\ref{tab:comparison_across_family}. We observe that more high-resource language families, e.g., Indo-European (\texttt{indo1319}) and Dravidian (\texttt{drav1251}), achieve slightly better performance than low-resource language families, e.g., Sino-Tibetan (\texttt{sino1245}).

\begin{table}[ht]
\centering
\small
\begin{tabular}{cc|ccc}
\toprule
family & $\|$Sent$\|$ & PBC (\nll$\downarrow$) & \sib (ACC $\uparrow$) & \taxi(ACC $\uparrow$)\\
\midrule
indo1319 & 988M & 145.35 & 63.53 & 53.03 \\
drav1251 & 135M & \textbf{131.29} & 56.25 & \textbf{54.65} \\
aust1307 & 113M & 147.37 & 62.83 & 49.69 \\
turk1311 & 109M & 161.71 & 57.08 & 52.55 \\
afro1255 & 100M & 165.46 & 52.00 & 43.74 \\
atla1278 & 57M & 141.92 & 42.90 & 45.52 \\
ural1272 & 50M & 137.52 & \textbf{66.67} & 48.58 \\
sino1245 & 29M & 155.64 & 49.30 & 49.31 \\
other & 60M & 167.69 & 55.74 & 46.67 \\
\bottomrule
\end{tabular}
\caption{Performance comparison across language families on PBC, \sib, and \taxi. $\|$Sent$\|$: sentence number used for continued pretraining in total. $\downarrow$ indicates the lower, the better. $\uparrow$ indicates the higher, the better.}
\label{tab:comparison_across_family}
\end{table}

In \Cref{tab:lang_cmp}, we present a comprehensive analysis of the top 5 performance improvements and declines across languages on \sib from \modelname compared to LLaMA 2-7B. We observe that \modelname has substantial improvements on low-resource scripts, e.g., Kannada (\texttt{kan\_Knda}), while has worse performance on high-resource languages, e.g., Swedish (\texttt{swe\_Latn}), which have been well covered by LLaMA 2-7B.

\begin{table}[ht]
    \centering
    \resizebox{\textwidth}{!}{
\begin{tabular}{cccc||cccc}
    \toprule
    \multicolumn{4}{c||}{high end} & \multicolumn{4}{c}{low end} \\
    \midrule
    Language & LLaMA 2-7B & \modelname & $\Delta$ & Language & LLaMA 2-7B & \modelname & $\Delta$ \\
    \midrule
kan\_Knda & 17.16 & 57.35 & 40.19 & swe\_Latn & 71.08 & 60.29 & -10.79 \\
ckb\_Arab & 19.61 & 60.29 & 40.68 & rus\_Cyrl & 71.57 & 65.20 & -06.37 \\
asm\_Beng & 17.16 & 58.82 & 41.66 & dan\_Latn & 69.12 & 63.24 & -05.88 \\
pan\_Guru & 14.22 & 58.82 & 44.60 & pol\_Latn & 74.51 & 68.63 & -05.88 \\
sin\_Sinh & 15.20 & 60.29 & 45.09 & ukr\_Cyrl & 71.57 & 65.69 & -05.88 \\
\bottomrule
\end{tabular}
}
    \caption{Results for five languages each with the largest (high end) and smallest (low end) gains from \modelname vs. LLaMA 2-7B for \sib.
    $\Delta$ indicates the difference between the scores of \modelname and LLaMA 2-7B.  
    See \S\ref{sec:detailed_results} for detailed results for each task.}
    \label{tab:lang_cmp}
\end{table}

In our comprehensive analysis of contributing factors on \sib, we note that the corpus size of a language exhibits a weak correlation of 0.13 with its performance gain. In contrast, the corpus size of the language family to which a language belongs demonstrates a moderate correlation of 0.40. A moderately high Pearson correlation of 0.53 is observed between the effect of vocabulary extension, i.e., the reduction in segmentation length, and the performance gain. This observation holds true for languages with both non-Latin scripts, such as Kannada (\texttt{kan\_Knda}), Malayalam (\texttt{mal\_Mlym}), and Tigrinya (\texttt{tir\_Ethi}), as well as Latin scripts, such as Igbo (\texttt{ibo\_Latn}) and Yoruba (\texttt{yor\_Latn}). It demonstrates the effectiveness of vocabulary extension.

\subsection{Effect of Number of Shots}

\begin{figure}[ht]
    \centering
    \resizebox {0.6\columnwidth} {!} {
    \includegraphics[clip]{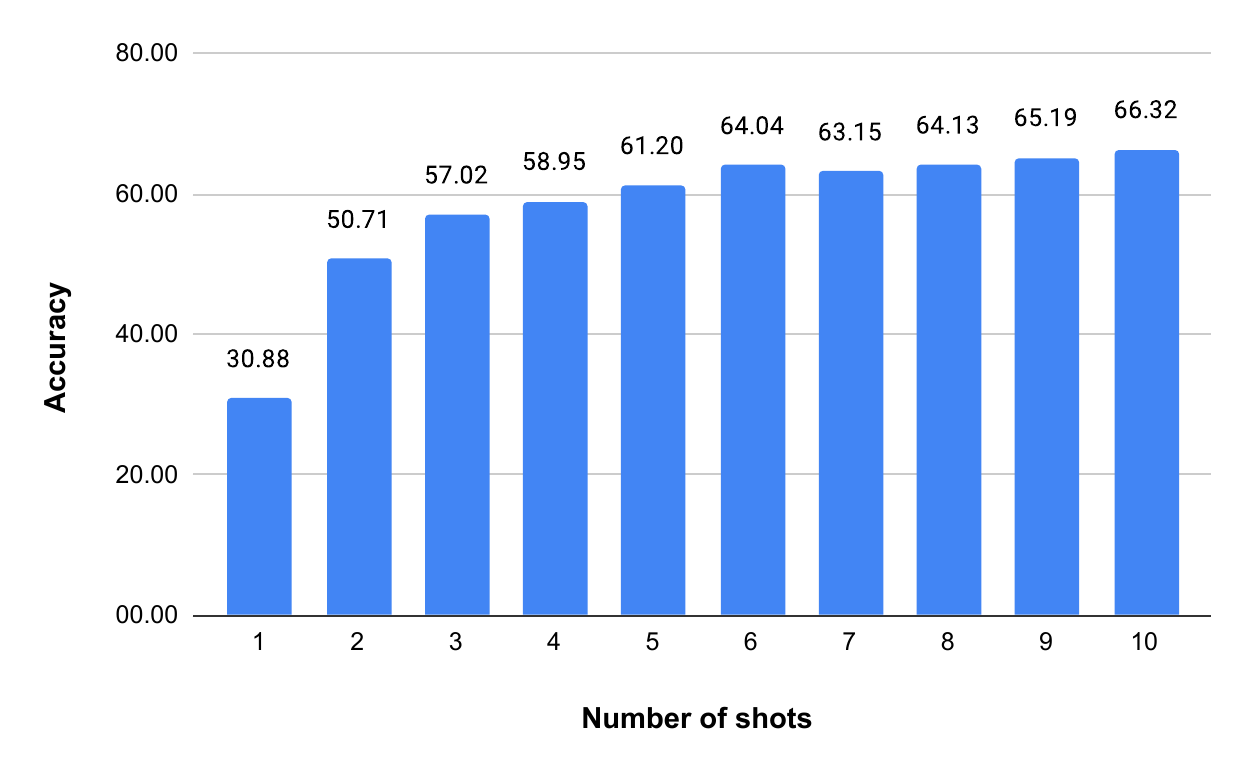}
    }
    \captionof{figure}{In-context learning macro-average accuracy (\%) on \sib with different number of shots using \modelname.}
    \label{fig:shot_effect_main}
\end{figure}

\Cref{fig:shot_effect_main} illustrates the relationship between accuracy and the number of in-context examples (i.e., shots) on \sib. As the number of in-context shots increases, there is a corresponding rise in accuracy. Notably, with just 1-shot, accuracy exhibits randomness at 30.88\%, indicating 1-shot provides limited information for task learning. The transition from 1 shot to 2 shots/3 shots results in a notable improvement, with performances boosted by 19.83\% and 26.14\%, respectively. This highlights the effectiveness of increasing the number of shots. \modelname achieves its peak performance at approximately 65\% accuracy with 6-10 in-context shots. This may be attributed to the multi-class nature of the \sib dataset, necessitating more shots for learning intricate input-label mappings.

\begin{figure}[ht]
    \centering
    \resizebox {0.6\columnwidth} {!} {
    \includegraphics[clip]{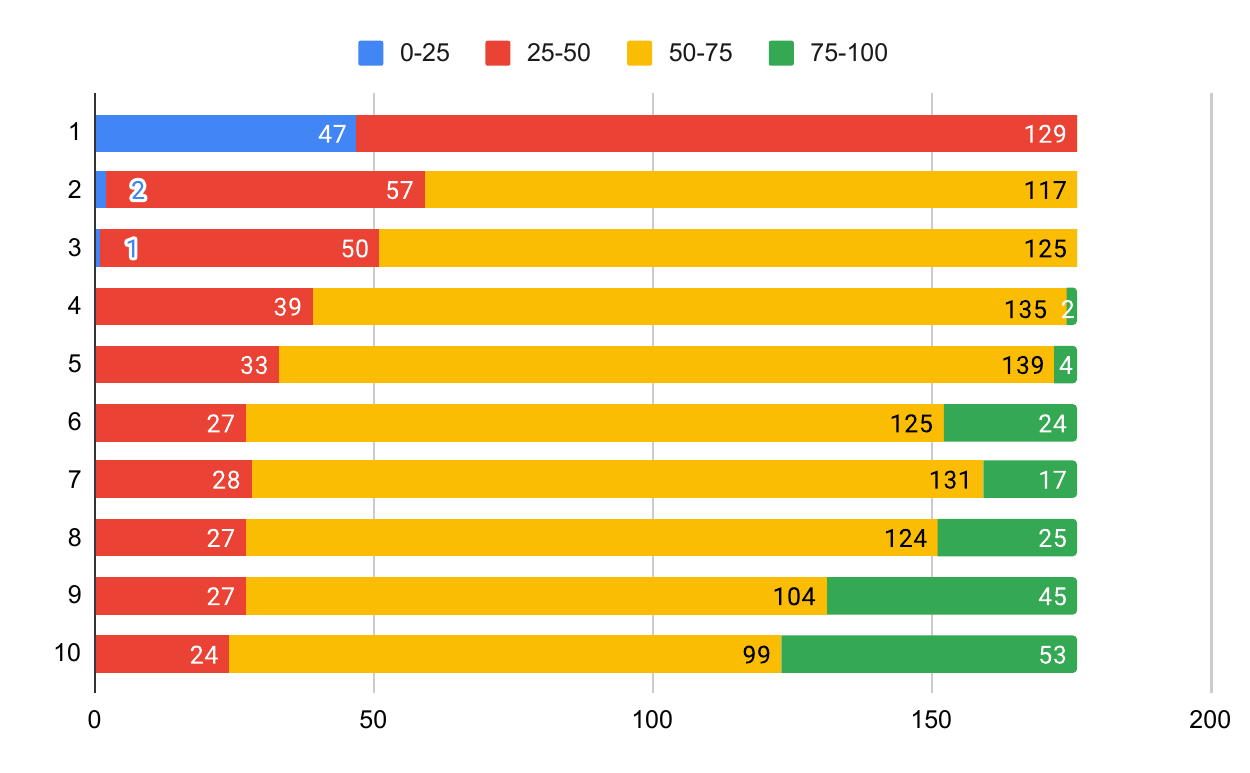}
    }
    \captionof{figure}{Detailed results of in-context learning on \sib using \modelname. X-axis: the number of languages in different accuracy ranges (\%). Y-axis: number of shots.}
    \label{fig:shot_effect_detail}
\end{figure}

In \Cref{fig:shot_effect_detail}, a more nuanced portrayal of results aligns with the observations made in \Cref{fig:shot_effect_main}. In the realm of 1-shot in-context learning, approximately 50 languages exhibit erratic results. As the number of shots increases, there is a reduction in the number of languages achieving low accuracy (25-50\%), coupled with a growing cohort achieving high accuracy (75-100\%). %

Further examination into individual language trends reveals that some low-resource languages require more shots to achieve better performance (e.g., \texttt{pes\_Arab} for Persian) or even exhibit poor performance with 10 shots (e.g., \texttt{dzo\_Tibt} for Dzongkha and \texttt{ayr\_Latn} for Central Aymara). In contrast, high-resource languages, such as \texttt{fra\_Latn} for French, demonstrate impressive performance even with fewer shots, and increasing the number of shots results in only marginal improvement.

\section{Related Work}

\subsection{Multilingual Language Models}

Language model development has endeavored to broaden the scope of pretraining languages to address multilingual scenarios.
Pretrained multilingual models have been able to accommodate up to a hundred or more languages.
Noteworthy examples include mBERT~\cite{devlin2019bert}, which supports 104 languages, XLM-R~\citep{conneau2020unsupervised} covering 100 languages, mBART~\citep{liu2020multilingual} designed for 25 languages, mT5~\citep{xue2021mt5} spanning 101 languages, XGLM~\citep{DBLP:journals/corr/abs-2112-10668} across 30 languages, GPT-3 covering 118 languages (93\% English), mGPT~\citep{DBLP:journals/corr/abs-2204-07580} accommodating 60 languages, and BLOOM~\citep{scao2022bloom} supporting 46 languages and 13 programming languages. 

Surprisingly, two recent multilingual language models have surpassed the conventional limit by supporting more than 400 languages. Glot500-m~\citep{imanigooghari-etal-2023-glot500} spans 511 languages through vocabulary extension and continued training based on XLM-R. 
SERENGETI~\citep{adebara2022serengeti} goes even further by supporting 517 African languages and language varieties, written in five different scripts, employing models inspired by both ELECTRA~\citep{clark2020ELECTRA} and XLM-R. 
MADLAD~\citep{kudugunta2023madlad} covers 419 languages and trains an 8B language model from scratch with an adapted UL2 objective~\citep{tay2022ul2}.  
Our work is concurrent with the MADLAD-400 language model. 
We distinguish it by: 
1) language coverage. Our work covered more than 500 languages, a number comparable to that of encoder-only models and surpassing MADLAD-400 by an additional 100 languages.
2) training methods. We consider continual training to benefit from the learned knowledge of the original models.
3) model architecture. We adopt an open model architecture, i.e., LLaMA, while MADLAD uses decoder-only T5 architecture, which has not been supported by the HuggingFace ecosystem at the time of writing, thus leading to additional difficulty in usage.

\subsection{Language Adaptation}

Before the advent of LLMs, diverse approaches are employed to adapt small-scale multilingual language models to new languages. These methods include using adapters \citep{DBLP:conf/emnlp/PfeifferVGR20,DBLP:conf/emnlp/UstunBBN20,DBLP:conf/emnlp/PfeifferVGR20,DBLP:conf/eacl/NguyenLVN21,DBLP:journals/corr/abs-2205-09634,DBLP:journals/corr/abs-2212-09535}, vocabulary extension and substitution \citep{DBLP:conf/emnlp/ChauLS20,DBLP:conf/emnlp/WangK0R20,DBLP:journals/corr/abs-2005-00318,DBLP:conf/naacl/MullerASS21,DBLP:conf/emnlp/PfeifferVGR21,DBLP:journals/corr/abs-2307-01163,DBLP:journals/corr/abs-2309-04679}, leveraging monolingual corpora \citep{DBLP:conf/acl/EbrahimiK20,DBLP:conf/coling/AlabiAMK22}, and utilizing bilingual lexicons \citep{DBLP:conf/acl/WangRN22}.

While language models have been scaled up notably, their coverage is limited to a specific set of languages. To address this constraint, various methods have been proposed to expand the applicability of these large language models across a broader range of languages, catering to both general-purpose tasks and specific applications like machine translation. These methods also involve vocabulary extension \citep{DBLP:journals/corr/abs-2304-08177}, continued pretraining and instruction-tuning \citep{DBLP:journals/corr/abs-2212-09535,DBLP:journals/corr/abs-2304-08177,DBLP:journals/corr/abs-2309-08958,zhao2024llama}, and parallel corpora exploitation \citep{DBLP:journals/corr/abs-2305-13627, DBLP:journals/corr/abs-2305-18098,DBLP:journals/corr/abs-2308-04948,DBLP:journals/corr/abs-2309-11674}. Despite these efforts, massive language adaptation of LLMs for general-purpose tasks across diverse languages, e.g., covering many languages families and more than one hundred languages, remains an area yet to be thoroughly explored.

\section{Conclusion and Future Work}

We present a pioneering effort in massive language adaptation on LLMs,  focusing on extending LLaMA 7B to our model, \modelname. This adaptation involves vocabulary extension and continued pretraining with LoRA. Our approach leads to \modelname achieving state-of-the-art in-context learning capabilities, as demonstrated on the benchmarks of \sib and \taxi. 
We release the training scripts and model weights publicly to facilitate future research.
This work marks a substantial advancement in applying LLMs to a diverse range of languages.

Our future work will focus on further improving the model capacity, for example, on machine translation across many language pairs. 
\citet{alves2023steering} showed that LLMs (LLaMA-7B and LLaMA-13B) exhibited poor performance even on English-centric high-resource language pairs in some cases. 
Translation with LLMs on low-resource languages is more challenging. 
The LLaMA-7B model performed poorly in our preliminary experiments.
Besides, our pretraining corpus does not intentionally include bilingual texts, and our \modelname model is not instruction-tuned with translation data. 
We leave the inclusion of bilingual text during continual pretraining, instruction fine-tuning with translation data, and the evaluation on machine translation as future works.

\clearpage

\section*{Ethical Statement}
LLMs have been known to exhibit biases present in their training data. When extending LLMs to low-resource languages, there is a risk of propagating biases from high-resource languages to underrepresented ones. Careful attention must be paid to mitigate bias and ensure fairness in data collection and model training. 
The paper aims to make LLMs more accessible for underrepresented languages. Still, there is a risk of creating a digital language divide if certain communities are left out due to limited technological access. 
Future work would address biases by conducting bias audits on the training data, debiasing the models during generation, and continuously monitoring model outputs.

\section*{Reproducibility Statement}
We make the following efforts to ensure reproducible research. 
We release the model weights (\url{https://huggingface.co/MaLA-LM}) and codes for training and evaluation (\url{https://github.com/MaLA-LM/mala-500}). 
We use publicly available evaluation benchmarks which can be obtained freely or by request. 
The results are reproducible with our released model weights and evaluation scripts, 

\section*{Acknowledgements}

We thank José Pombal for constructive suggestions on training.
This work is funded by The European Research Council  (grants \#740516, \#771113 and \#758969), EU's Horizon Europe Research and Innovation Actions (UTTER, contract 101070631), and the European Union's Horizon Europe research and innovation programme under grant agreement No 101070350 and from UK Research and Innovation (UKRI) under the UK government’s Horizon Europe funding guarantee [grant \#10052546].
The authors wish to acknowledge CSC – IT Center for Science, Finland, for generous computational resources on the Mahti supercomputer and LUMI supercomputer through the LUMI extreme scale access (MOOMIN and LumiNMT).
Shaoxiong Ji and Peiqin Lin acknowledge travel support from ELISE (GA no 951847).

\bibliography{LLM500}
\bibliographystyle{colm2024_conference}

\appendix
\section{Languages}
\label{sec:appendix_languages}
The list of languages of Glot500-c used to train \modelname 
with the number of available sentences and language family information for each language is available 
in Tables~\ref{languages1}, ~\ref{languages2} and \ref{languages3}. %

\begin{table*}
\centering
\resizebox{0.85\textwidth}{!}{
\begin{tabular}{ccc|ccc|ccc}
    \toprule
    Lang & $\|Sent\|$ & Family & Lang & $\|Sent\|$ & Family & Lang & $\|Sent\|$ & Family \\
    \midrule
hbs\_Latn & 63411156 & indo1319 & hin\_Deva & 7046700 & indo1319 & ton\_Latn & 1216118 & aust1307 \\
mal\_Mlym & 48098273 & drav1251 & kor\_Hang & 6468444 & kore1284 & tah\_Latn & 1190747 & aust1307 \\
aze\_Latn & 46300705 & & ory\_Orya & 6266475 & indo1319 & lat\_Latn & 1179913 & indo1319 \\
guj\_Gujr & 45738685 & indo1319 & urd\_Arab & 6009594 & indo1319 & srn\_Latn & 1172349 & indo1319 \\
ben\_Beng & 43514870 & indo1319 & swa\_Latn & 5989369 & & ewe\_Latn & 1161605 & atla1278 \\
kan\_Knda & 41836495 & drav1251 & sqi\_Latn & 5526836 & indo1319 & bem\_Latn & 1111969 & atla1278 \\
tel\_Telu & 41580525 & drav1251 & bel\_Cyrl & 5319675 & indo1319 & efi\_Latn & 1082621 & atla1278 \\
\bottomrule
    \end{tabular}
}
    \caption{List of languages of Glot500-c (Part I).}
    \label{languages1}
\end{table*}

\begin{table*}
\centering
\resizebox{0.85\textwidth}{!}{
    \begin{tabular}{ccc|ccc|ccc}
    \toprule
    Lang & $\|Sent\|$ & Family & Lang & $\|Sent\|$ & Family & Lang & $\|Sent\|$ & Family \\
    \midrule
lus\_Latn & 509250 & sino1245 & kmb\_Latn & 296269 & atla1278 & ncx\_Latn & 162558 & utoa1244 \\
crs\_Latn & 508755 & indo1319 & zai\_Latn & 277632 & otom1299 & qug\_Latn & 162500 & quec1387 \\
kqn\_Latn & 507913 & atla1278 & gym\_Latn & 274512 & chib1249 & rmn\_Latn & 162069 & indo1319 \\
ndo\_Latn & 496613 & atla1278 & bod\_Tibt & 273489 & sino1245 & cjk\_Latn & 160645 & atla1278 \\
snd\_Arab & 488730 & indo1319 & nde\_Latn & 269931 & atla1278 & arb\_Arab & 159884 & afro1255 \\
yue\_Hani & 484700 & sino1245 & fon\_Latn & 268566 & atla1278 & kea\_Latn & 158047 & indo1319 \\
tiv\_Latn & 483064 & atla1278 & ber\_Latn & 264426 & & mck\_Latn & 157521 & atla1278 \\
\bottomrule
    \end{tabular}
}
    \caption{List of languages of Glot500-c (Part II).}
    \label{languages2}
\end{table*}

\begin{table*}
\centering
\resizebox{0.85\textwidth}{!}{
    \begin{tabular}{ccc|ccc|ccc}
    \toprule
    Lang & $\|Sent\|$ & Family & Lang & $\|Sent\|$ & Family & Lang & $\|Sent\|$ & Family \\
    \midrule
alt\_Cyrl & 95148 & turk1311 & mny\_Latn & 50581 & atla1278 & csy\_Latn & 34126 & sino1245 \\
rmn\_Grek & 94533 & indo1319 & gkp\_Latn & 50549 & mand1469 & azb\_Arab & 33758 & turk1311 \\
miq\_Latn & 94343 & misu1242 & kat\_Latn & 50424 & kart1248 & csb\_Latn & 33743 & indo1319 \\
kaa\_Cyrl & 88815 & turk1311 & bjn\_Latn & 49068 & aust1307 & tpm\_Latn & 33517 & atla1278 \\
kos\_Latn & 88603 & aust1307 & acr\_Latn & 48886 & maya1287 & quw\_Latn & 33449 & quec1387 \\
grn\_Latn & 87568 & & dtp\_Latn & 48468 & aust1307 & rmy\_Cyrl & 33351 & indo1319 \\
lhu\_Latn & 87255 & sino1245 & lam\_Latn & 46853 & atla1278 & ixl\_Latn & 33289 & maya1287 \\
\bottomrule
    \end{tabular}
}
    \caption{List of languages of Glot500-c (Part III).}
    \label{languages3}
\end{table*}

\section{Detailed Results}
\label{sec:detailed_results}

Detailed results of evaluation are shown 
in Tables~\ref{pplm1}-\ref{pplm8} (\nll on Glot500-c), Tables~\ref{ppll1}-\ref{ppll6} (\nll on PBC), Tables~\ref{result1}-\ref{result2} (ACC on \sib), and Tables~\ref{taxi1}-\ref{taxi6} (ACC on \taxi).

\begin{table*}
\centering
\resizebox{0.81\textwidth}{!}{
\begin{tabular}{c|ccccc}
\toprule
Lang & LLaMA 2-7B & mGPT-13B & BLOOM-7B1 & XGLM-7.5B & MALA-500 \\
\midrule
abk\_Cyrl & 234.09 & 249.16 & 258.26 & 231.44 & 164.61 \\
abn\_Latn & 140.01 & 197.81 & 153.58 & 152.90 & 111.86 \\
ace\_Latn & 235.15 & 332.18 & 244.00 & 259.64 & 168.79 \\
ach\_Latn & 179.03 & 227.84 & 194.55 & 197.05 & 161.01 \\
acm\_Arab & 119.15 & 153.09 & 106.29 & 101.35 & 135.82 \\
acr\_Latn & 301.73 & 399.80 & 321.79 & 316.49 & 194.71 \\
ada\_Latn & 132.76 & 168.56 & 150.19 & 137.99 & 103.17 \\
\bottomrule
\end{tabular}
}
\caption{Detailed results of \nll on Glot500-c (Part I).}
\label{pplm1}
\end{table*}

\begin{table*}
\centering
\resizebox{0.81\textwidth}{!}{
\begin{tabular}{c|ccccc}
\toprule
Lang & LLaMA 2-7B & mGPT-13B & BLOOM-7B1 & XGLM-7.5B & MALA-500 \\
\midrule
cab\_Latn & 222.05 & 292.04 & 234.53 & 237.57 & 168.63 \\
cac\_Latn & 293.47 & 395.52 & 310.33 & 301.30 & 192.22 \\
cak\_Latn & 295.24 & 394.87 & 317.52 & 309.03 & 200.69 \\
caq\_Latn & 240.00 & 323.71 & 264.17 & 257.49 & 164.95 \\
cat\_Latn & 94.68 & 212.17 & 83.26 & 86.26 & 130.00 \\
cbk\_Latn & 143.05 & 221.60 & 145.69 & 159.41 & 137.96 \\
cce\_Latn & 178.45 & 226.07 & 190.01 & 192.54 & 152.70 \\
\bottomrule
\end{tabular}
}
\caption{Detailed results of \nll on Glot500-c (Part II).}
\label{pplm2}
\end{table*}

\begin{table*}
\centering
\resizebox{0.81\textwidth}{!}{
\begin{tabular}{c|ccccc}
\toprule
Lang & LLaMA 2-7B & mGPT-13B & BLOOM-7B1 & XGLM-7.5B & MALA-500 \\
\midrule
gkn\_Latn & 223.58 & 304.46 & 253.54 & 245.24 & 167.81 \\
gkp\_Latn & 261.56 & 358.97 & 280.80 & 270.48 & 186.41 \\
gla\_Latn & 220.92 & 382.20 & 293.89 & 315.23 & 210.51 \\
gle\_Latn & 203.10 & 345.45 & 276.11 & 299.80 & 206.52 \\
glg\_Latn & 120.88 & 204.76 & 108.43 & 122.45 & 132.58 \\
glv\_Latn & 232.86 & 326.69 & 247.79 & 265.04 & 182.93 \\
gom\_Deva & 328.82 & 462.17 & 324.77 & 358.50 & 233.15 \\
\bottomrule
\end{tabular}
}
\caption{Detailed results of \nll on Glot500-c (Part III).}
\label{pplm3}
\end{table*}

\begin{table*}
\centering
\resizebox{0.81\textwidth}{!}{
\begin{tabular}{c|ccccc}
\toprule
Lang & LLaMA 2-7B & mGPT-13B & BLOOM-7B1 & XGLM-7.5B & MALA-500 \\
\midrule
jav\_Latn & 208.92 & 275.29 & 212.60 & 220.31 & 180.00 \\
jbo\_Latn & 103.91 & 200.82 & 109.86 & 112.61 & 117.25 \\
jpn\_Jpan & 136.26 & 301.32 & 197.23 & 149.70 & 150.43 \\
kaa\_Cyrl & 281.21 & 363.07 & 300.20 & 317.13 & 146.98 \\
kaa\_Latn & 284.60 & 354.51 & 292.04 & 309.67 & 192.43 \\
kab\_Latn & 192.58 & 264.51 & 185.56 & 216.46 & 161.31 \\
kac\_Latn & 210.47 & 267.38 & 223.77 & 249.95 & 166.44 \\
\bottomrule
\end{tabular}
}
\caption{Detailed results of \nll on Glot500-c (Part IV).}
\label{pplm4}
\end{table*}

\begin{table*}
\centering
\resizebox{0.81\textwidth}{!}{
\begin{tabular}{c|ccccc}
\toprule
Lang & LLaMA 2-7B & mGPT-13B & BLOOM-7B1 & XGLM-7.5B & MALA-500 \\
\midrule
luo\_Latn & 177.43 & 224.04 & 194.07 & 187.72 & 156.23 \\
lus\_Latn & 192.97 & 251.35 & 203.37 & 212.72 & 163.95 \\
lvs\_Latn & 154.85 & 211.40 & 185.87 & 198.99 & 138.63 \\
lzh\_Hani & 149.57 & 215.19 & 130.32 & 153.38 & 151.15 \\
mad\_Latn & 232.71 & 325.38 & 245.39 & 249.29 & 176.81 \\
mah\_Latn & 178.50 & 246.26 & 188.98 & 183.17 & 145.35 \\
mai\_Deva & 245.94 & 389.84 & 189.93 & 223.00 & 185.84 \\
\bottomrule
\end{tabular}
}
\caption{Detailed results of \nll on Glot500-c (Part V).}
\label{pplm5}
\end{table*}

\begin{table*}
\centering
\resizebox{0.81\textwidth}{!}{
\begin{tabular}{c|ccccc}
\toprule
Lang & LLaMA 2-7B & mGPT-13B & BLOOM-7B1 & XGLM-7.5B & MALA-500 \\
\midrule
nse\_Latn & 176.20 & 234.62 & 184.77 & 174.57 & 161.52 \\
nso\_Latn & 170.49 & 227.97 & 170.96 & 201.29 & 142.72 \\
nya\_Latn & 203.45 & 299.12 & 222.51 & 224.89 & 175.69 \\
nyk\_Latn & 131.13 & 166.47 & 142.89 & 138.50 & 105.26 \\
nyn\_Latn & 174.51 & 229.51 & 189.05 & 194.14 & 149.17 \\
nyu\_Latn & 126.29 & 172.40 & 132.49 & 127.67 & 99.26 \\
nyy\_Latn & 215.07 & 271.23 & 234.37 & 220.80 & 168.74 \\
\bottomrule
\end{tabular}
}
\caption{Detailed results of \nll on Glot500-c (Part VI).}
\label{pplm6}
\end{table*}

\begin{table*}
\centering
\resizebox{0.81\textwidth}{!}{
\begin{tabular}{c|ccccc}
\toprule
Lang & LLaMA 2-7B & mGPT-13B & BLOOM-7B1 & XGLM-7.5B & MALA-500 \\
\midrule
sat\_Olck & 654.37 & 3377.97 & 667.66 & 40.17 & 311.96 \\
sba\_Latn & 272.45 & 372.47 & 303.48 & 293.62 & 167.13 \\
scn\_Latn & 236.20 & 355.24 & 263.10 & 270.02 & 191.69 \\
sco\_Latn & 147.94 & 341.79 & 193.24 & 193.20 & 170.39 \\
seh\_Latn & 173.46 & 231.41 & 177.38 & 174.40 & 138.70 \\
sgs\_Latn & 248.33 & 313.78 & 251.35 & 277.73 & 182.16 \\
sid\_Latn & 135.53 & 180.29 & 147.72 & 139.17 & 114.24 \\
\bottomrule
\end{tabular}
}
\caption{Detailed results of \nll on Glot500-c (Part VII).}
\label{pplm7}
\end{table*}

\begin{table*}
\centering
\resizebox{0.81\textwidth}{!}{
\begin{tabular}{c|ccccc}
\toprule
Lang & LLaMA 2-7B & mGPT-13B & BLOOM-7B1 & XGLM-7.5B & MALA-500 \\
\midrule
tsz\_Latn & 183.82 & 253.97 & 200.16 & 176.35 & 153.98 \\
ttj\_Latn & 133.35 & 174.08 & 142.10 & 146.60 & 112.98 \\
tuc\_Latn & 325.34 & 444.23 & 346.52 & 291.84 & 180.84 \\
tui\_Latn & 247.40 & 330.20 & 266.54 & 265.71 & 181.39 \\
tuk\_Cyrl & 196.40 & 248.01 & 210.45 & 219.39 & 143.19 \\
tuk\_Latn & 217.31 & 235.95 & 217.78 & 238.66 & 155.71 \\
tum\_Latn & 184.51 & 236.91 & 190.41 & 153.36 & 153.44 \\
\bottomrule
\end{tabular}
}
\caption{Detailed results of \nll on Glot500-c (Part VIII).}
\label{pplm8}
\end{table*}

\begin{table*}
\centering
\resizebox{0.81\textwidth}{!}{
\begin{tabular}{c|ccccc}
\toprule
Lang & LLaMA 2-7B & mGPT-13B & BLOOM-7B1 & XGLM-7.5B & MALA-500 \\
\midrule
ace\_Latn & 137.43 & 196.93 & 144.50 & 152.49 & 97.89 \\
ach\_Latn & 113.66 & 152.08 & 123.29 & 125.31 & 102.45 \\
acr\_Latn & 177.86 & 233.22 & 188.27 & 182.33 & 114.27 \\
afr\_Latn & 80.43 & 132.25 & 116.34 & 129.21 & 95.33 \\
agw\_Latn & 130.32 & 186.58 & 136.17 & 138.27 & 95.93 \\
ahk\_Latn & 175.75 & 291.31 & 187.63 & 179.54 & 116.76 \\
aka\_Latn & 98.41 & 135.74 & 99.46 & 108.01 & 78.20 \\
\bottomrule
\end{tabular}
}
\caption{Detailed results of \nll on PBC (Part I).}
\label{ppll1}
\end{table*}

\begin{table*}
\centering
\resizebox{0.81\textwidth}{!}{
\begin{tabular}{c|ccccc}
\toprule
Lang & LLaMA 2-7B & mGPT-13B & BLOOM-7B1 & XGLM-7.5B & MALA-500 \\
\midrule
dan\_Latn & 63.65 & 114.43 & 97.95 & 101.13 & 86.68 \\
deu\_Latn & 57.09 & 109.69 & 84.08 & 54.00 & 80.90 \\
djk\_Latn & 143.19 & 192.80 & 147.13 & 153.74 & 120.66 \\
dln\_Latn & 113.43 & 155.19 & 118.82 & 125.73 & 92.37 \\
dtp\_Latn & 158.44 & 222.01 & 165.46 & 169.63 & 111.77 \\
dyu\_Latn & 122.24 & 161.61 & 126.53 & 132.73 & 103.04 \\
dzo\_Tibt & 157.37 & 550.44 & 162.42 & 36.35 & 99.22 \\
\end{tabular}
}
\caption{Detailed results of \nll on PBC (Part II).}
\label{ppll2}
\end{table*}

\begin{table*}
\centering
\resizebox{0.81\textwidth}{!}{
\begin{tabular}{c|ccccc}
\toprule
Lang & LLaMA 2-7B & mGPT-13B & BLOOM-7B1 & XGLM-7.5B & MALA-500 \\
\midrule
ium\_Latn & 150.62 & 222.39 & 155.20 & 157.52 & 99.54 \\
ixl\_Latn & 190.07 & 299.20 & 206.08 & 202.92 & 127.52 \\
izz\_Latn & 167.28 & 228.45 & 195.19 & 198.57 & 118.78 \\
jam\_Latn & 119.85 & 181.93 & 134.52 & 139.07 & 96.42 \\
jav\_Latn & 134.11 & 171.16 & 136.06 & 140.01 & 109.34 \\
jpn\_Jpan & 67.67 & 114.11 & 84.64 & 61.57 & 88.53 \\
kaa\_Cyrl & 136.14 & 179.48 & 138.63 & 153.33 & 84.79 \\
\bottomrule
\end{tabular}
}
\caption{Detailed results of \nll on PBC (Part III).}
\label{ppll3}
\end{table*}

\begin{table*}
\centering
\resizebox{0.81\textwidth}{!}{
\begin{tabular}{c|ccccc}
\toprule
Lang & LLaMA 2-7B & mGPT-13B & BLOOM-7B1 & XGLM-7.5B & MALA-500 \\
\midrule
mhr\_Cyrl & 122.42 & 160.48 & 119.36 & 127.42 & 100.09 \\
min\_Latn & 139.41 & 194.79 & 136.22 & 138.87 & 133.30 \\
miq\_Latn & 129.28 & 182.98 & 144.36 & 141.12 & 104.92 \\
mkd\_Cyrl & 85.29 & 151.22 & 112.67 & 89.21 & 89.46 \\
mlg\_Latn & 107.66 & 135.73 & 106.88 & 128.75 & 86.60 \\
mlt\_Latn & 108.58 & 168.92 & 134.24 & 137.26 & 107.12 \\
mos\_Latn & 129.97 & 161.07 & 135.30 & 138.61 & 112.98 \\
\bottomrule
\end{tabular}
}
\caption{Detailed results of \nll on PBC (Part IV).}
\label{ppll4}
\end{table*}

\begin{table*}
\centering
\resizebox{0.81\textwidth}{!}{
\begin{tabular}{c|ccccc}
\toprule
Lang & LLaMA 2-7B & mGPT-13B & BLOOM-7B1 & XGLM-7.5B & MALA-500 \\
\midrule
qug\_Latn & 124.11 & 158.66 & 131.62 & 95.23 & 95.07 \\
quh\_Latn & 148.83 & 174.81 & 154.34 & 107.04 & 106.24 \\
quw\_Latn & 104.78 & 139.63 & 109.91 & 95.69 & 92.58 \\
quy\_Latn & 119.84 & 140.49 & 127.93 & 85.16 & 94.14 \\
quz\_Latn & 126.18 & 149.08 & 134.60 & 85.68 & 96.32 \\
qvi\_Latn & 134.03 & 177.51 & 139.73 & 114.64 & 100.81 \\
rap\_Latn & 139.27 & 239.23 & 152.39 & 157.18 & 100.81 \\
\bottomrule
\end{tabular}
}
\caption{Detailed results of \nll on PBC (Part V).}
\label{ppll5}
\end{table*}

\begin{table*}
\centering
\resizebox{0.81\textwidth}{!}{
\begin{tabular}{c|ccccc}
\toprule
Lang & LLaMA 2-7B & mGPT-13B & BLOOM-7B1 & XGLM-7.5B & MALA-500 \\
\midrule
tui\_Latn & 135.41 & 187.71 & 146.21 & 146.43 & 107.57 \\
tuk\_Cyrl & 127.20 & 168.02 & 136.69 & 145.86 & 94.72 \\
tuk\_Latn & 123.72 & 144.22 & 124.67 & 135.21 & 97.42 \\
tum\_Latn & 127.49 & 165.45 & 130.05 & 109.19 & 102.35 \\
tur\_Latn & 76.97 & 118.11 & 102.96 & 57.95 & 99.22 \\
twi\_Latn & 110.21 & 159.12 & 110.54 & 122.43 & 93.81 \\
tyv\_Cyrl & 165.82 & 197.37 & 164.25 & 181.87 & 107.33 \\
\bottomrule
\end{tabular}
}
\caption{Detailed results of \nll on PBC (Part VI).}
\label{ppll6}
\end{table*}

\begin{table*}
\centering
\resizebox{0.81\textwidth}{!}{
\begin{tabular}{c|cccc|cccccccccc}
\toprule
\multirow{2}{*}{Lang} & \multirow{2}{*}{\makecell{LLaMA 2 \\ 7B}} & \multirow{2}{*}{\makecell{mGPT \\ 13B}} & \multirow{2}{*}{\makecell{BLOOM \\ 7B1}} & \multirow{2}{*}{\makecell{XGLM \\ 7.5B}} & \multicolumn{10}{c}{\modelname} \\
& & & & & 1-shot & 2-shot & 3-shot & 4-shot & 5-shot & 6-shot & 7-shot & 8-shot & 9-shot & 10-shot \\
\midrule
ace\_Latn & 44.12 & 47.55 & 50.00 & 36.76 & 34.31 & 52.94 & 60.29 & 60.78 & 65.69 & 67.65 & 64.22 & 65.20 & 68.63 & 71.57 \\
acm\_Arab & 52.45 & 65.69 & 69.12 & 58.33 & 32.35 & 53.43 & 59.31 & 63.73 & 63.73 & 67.16 & 66.67 & 69.12 & 66.67 & 66.67 \\
afr\_Latn & 68.14 & 55.39 & 53.92 & 40.20 & 41.18 & 62.25 & 65.69 & 69.12 & 71.08 & 74.02 & 73.53 & 74.51 & 76.96 & 78.92 \\
ajp\_Arab & 47.55 & 64.22 & 68.63 & 53.43 & 33.33 & 56.86 & 59.80 & 59.80 & 65.20 & 63.73 & 63.24 & 69.12 & 68.14 & 66.67 \\
als\_Latn & 41.67 & 46.57 & 45.59 & 28.43 & 27.94 & 51.96 & 63.73 & 62.25 & 69.12 & 71.08 & 69.61 & 72.06 & 75.98 & 77.45 \\
amh\_Ethi & 15.69 & 18.63 & 16.67 & 13.24 & 25.00 & 36.76 & 45.59 & 51.47 & 51.96 & 53.92 & 51.96 & 53.43 & 54.90 & 53.92 \\
\bottomrule
\end{tabular}
}
\caption{Detailed results on \sib (Part I). For previous LLMs, 3-shot results are presented.}
\label{result1}
\end{table*}

\begin{table*}
\centering
\resizebox{0.81\textwidth}{!}{
\begin{tabular}{c|cccc|cccccccccc}
\toprule
\multirow{2}{*}{Lang} & \multirow{2}{*}{\makecell{LLaMA 2 \\ 7B}} & \multirow{2}{*}{\makecell{mGPT \\ 13B}} & \multirow{2}{*}{\makecell{BLOOM \\ 7B1}} & \multirow{2}{*}{\makecell{XGLM \\ 7.5B}} & \multicolumn{10}{c}{\modelname} \\
& & & & & 1-shot & 2-shot & 3-shot & 4-shot & 5-shot & 6-shot & 7-shot & 8-shot & 9-shot & 10-shot \\
\midrule
lim\_Latn & 60.78 & 50.00 & 50.00 & 33.82 & 32.84 & 56.37 & 60.78 & 63.24 & 66.67 & 67.16 & 69.12 & 72.55 & 70.59 & 72.06 \\
lin\_Latn & 36.76 & 40.20 & 43.14 & 34.31 & 23.04 & 38.24 & 47.06 & 53.92 & 57.84 & 61.27 & 56.86 & 60.29 & 62.75 & 65.69 \\
lit\_Latn & 40.20 & 60.29 & 41.18 & 30.39 & 32.35 & 55.39 & 62.75 & 65.20 & 64.71 & 70.59 & 68.14 & 66.67 & 69.61 & 73.04 \\
lmo\_Latn & 57.84 & 50.98 & 55.88 & 41.67 & 34.80 & 59.31 & 65.69 & 66.18 & 70.10 & 71.57 & 70.59 & 70.59 & 75.00 & 75.49 \\
ltz\_Latn & 55.88 & 47.06 & 52.94 & 39.22 & 39.22 & 56.37 & 65.20 & 61.27 & 70.59 & 68.14 & 71.08 & 70.59 & 71.08 & 74.02 \\
lua\_Latn & 32.35 & 33.33 & 39.22 & 28.43 & 20.10 & 33.82 & 40.20 & 42.65 & 49.02 & 51.96 & 50.00 & 50.00 & 49.51 & 50.00 \\
\bottomrule
\end{tabular}
}
\caption{Detailed results on \sib (Part II). For previous LLMs, 3-shot results are presented.}
\label{result2}
\end{table*}

\begin{table*}
\centering
\resizebox{0.81\textwidth}{!}{
\begin{tabular}{c|ccccc}
\toprule
Lang & LLaMA 2-7B & mGPT-13B & BLOOM-7B1 & XGLM-7.5B & \modelname \\
\midrule
ace\_Latn & 46.85 & 47.75 & 49.55 & 41.44 & 48.65 \\
ach\_Latn & 45.05 & 37.84 & 41.44 & 40.54 & 36.04 \\
acr\_Latn & 47.75 & 51.35 & 50.45 & 47.75 & 50.45 \\
afr\_Latn & 54.05 & 38.74 & 51.35 & 49.55 & 55.86 \\
agw\_Latn & 48.65 & 45.05 & 41.44 & 42.34 & 49.55 \\
ahk\_Latn & 43.24 & 36.04 & 36.04 & 35.14 & 45.05 \\
aka\_Latn & 42.34 & 32.43 & 38.74 & 42.34 & 54.95 \\
\bottomrule
\end{tabular}
}
\caption{Detailed results on \taxi (Part I). 3-shot results are presented.}
\label{taxi1}
\end{table*}

\begin{table*}
\centering
\resizebox{0.81\textwidth}{!}{
\begin{tabular}{c|ccccc}
\toprule
Lang & LLaMA 2-7B & mGPT-13B & BLOOM-7B1 & XGLM-7.5B & \modelname \\
\midrule
dtp\_Latn & 39.64 & 35.14 & 42.34 & 46.85 & 50.45 \\
dyu\_Latn & 41.44 & 39.64 & 42.34 & 38.74 & 46.85 \\
dzo\_Tibt & 45.05 & 40.54 & 41.44 & 45.05 & 45.05 \\
efi\_Latn & 39.64 & 36.04 & 38.74 & 41.44 & 45.95 \\
ell\_Grek & 49.55 & 45.95 & 49.55 & 48.65 & 51.35 \\
eng\_Latn & 55.86 & 42.34 & 58.56 & 54.05 & 59.46 \\
enm\_Latn & 50.45 & 41.44 & 56.76 & 50.45 & 54.95 \\
\bottomrule
\end{tabular}
}
\caption{Detailed results on \taxi (Part II). 3-shot results are presented.}
\label{taxi2}
\end{table*}

\begin{table*}
\centering
\resizebox{0.81\textwidth}{!}{
\begin{tabular}{c|ccccc}
\toprule
Lang & LLaMA 2-7B & mGPT-13B & BLOOM-7B1 & XGLM-7.5B & \modelname \\
\midrule
jpn\_Jpan & 46.85 & 46.85 & 47.75 & 50.45 & 51.35 \\
kaa\_Latn & 43.24 & 53.15 & 47.75 & 51.35 & 54.05 \\
kab\_Latn & 27.93 & 36.04 & 30.63 & 34.23 & 35.14 \\
kac\_Latn & 44.14 & 34.23 & 43.24 & 42.34 & 52.25 \\
kal\_Latn & 41.44 & 37.84 & 36.04 & 35.14 & 40.54 \\
kan\_Knda & 48.65 & 37.84 & 52.25 & 45.95 & 54.05 \\
kat\_Geor & 41.44 & 41.44 & 42.34 & 46.85 & 48.65 \\
\bottomrule
\end{tabular}
}
\caption{Detailed results on \taxi (Part III). 3-shot results are presented.}
\label{taxi3}
\end{table*}

\begin{table*}
\centering
\resizebox{0.81\textwidth}{!}{
\begin{tabular}{c|ccccc}
\toprule
Lang & LLaMA 2-7B & mGPT-13B & BLOOM-7B1 & XGLM-7.5B & \modelname \\
\midrule
mrw\_Latn & 40.54 & 39.64 & 41.44 & 37.84 & 49.55 \\
msa\_Latn & 44.14 & 41.44 & 45.95 & 37.84 & 46.85 \\
mwm\_Latn & 36.94 & 31.53 & 39.64 & 38.74 & 47.75 \\
mxv\_Latn & 33.33 & 35.14 & 39.64 & 38.74 & 40.54 \\
mya\_Mymr & 45.05 & 48.65 & 44.14 & 44.14 & 46.85 \\
myv\_Cyrl & 39.64 & 43.24 & 40.54 & 41.44 & 45.05 \\
mzh\_Latn & 45.05 & 45.95 & 42.34 & 40.54 & 44.14 \\
\bottomrule
\end{tabular}
}
\caption{Detailed results on \taxi (Part IV). 3-shot results are presented.}
\label{taxi4}
\end{table*}

\begin{table*}
\centering
\resizebox{0.81\textwidth}{!}{
\begin{tabular}{c|ccccc}
\toprule
Lang & LLaMA 2-7B & mGPT-13B & BLOOM-7B1 & XGLM-7.5B & \modelname \\
\midrule
rop\_Latn & 45.95 & 45.05 & 42.34 & 42.34 & 55.86 \\
rug\_Latn & 43.24 & 38.74 & 46.85 & 44.14 & 45.05 \\
run\_Latn & 46.85 & 40.54 & 45.05 & 40.54 & 52.25 \\
rus\_Cyrl & 49.55 & 41.44 & 50.45 & 47.75 & 53.15 \\
sag\_Latn & 43.24 & 43.24 & 41.44 & 40.54 & 47.75 \\
sah\_Cyrl & 40.54 & 35.14 & 44.14 & 44.14 & 54.95 \\
sba\_Latn & 42.34 & 43.24 & 45.05 & 40.54 & 49.55 \\
\bottomrule
\end{tabular}
}
\caption{Detailed results on \taxi (Part V). 3-shot results are presented.}
\label{taxi5}
\end{table*}

\begin{table*}
\centering
\resizebox{0.81\textwidth}{!}{
\begin{tabular}{c|ccccc}
\toprule
Lang & LLaMA 2-7B & mGPT-13B & BLOOM-7B1 & XGLM-7.5B & \modelname \\
\midrule
ukr\_Cyrl & 52.25 & 48.65 & 51.35 & 55.86 & 53.15 \\
uzb\_Latn & 45.05 & 49.55 & 37.84 & 46.85 & 54.05 \\
uzn\_Cyrl & 45.95 & 40.54 & 45.05 & 45.05 & 49.55 \\
ven\_Latn & 45.05 & 44.14 & 42.34 & 41.44 & 54.05 \\
vie\_Latn & 53.15 & 45.95 & 62.16 & 45.95 & 54.95 \\
wal\_Latn & 35.14 & 33.33 & 35.14 & 35.14 & 39.64 \\
war\_Latn & 48.65 & 39.64 & 37.84 & 45.05 & 54.95 \\
wbm\_Latn & 48.65 & 39.64 & 46.85 & 46.85 & 48.65 \\
wol\_Latn & 36.04 & 34.23 & 32.43 & 34.23 & 36.94 \\
xav\_Latn & 50.45 & 33.33 & 46.85 & 44.14 & 45.95 \\
xho\_Latn & 43.24 & 37.84 & 40.54 & 39.64 & 46.85 \\
yan\_Latn & 45.05 & 46.85 & 52.25 & 41.44 & 53.15 \\
yao\_Latn & 42.34 & 41.44 & 43.24 & 44.14 & 48.65 \\
yap\_Latn & 38.74 & 40.54 & 35.14 & 32.43 & 41.44 \\
yom\_Latn & 35.14 & 31.53 & 33.33 & 25.23 & 36.94 \\
yor\_Latn & 41.44 & 38.74 & 39.64 & 44.14 & 47.75 \\
yua\_Latn & 41.44 & 32.43 & 43.24 & 41.44 & 36.04 \\
yue\_Hani & 43.24 & 48.65 & 53.15 & 38.74 & 57.66 \\
zai\_Latn & 45.05 & 35.14 & 40.54 & 43.24 & 44.14 \\
zho\_Hani & 47.75 & 51.35 & 51.35 & 44.14 & 58.56 \\
zlm\_Latn & 54.05 & 49.55 & 57.66 & 56.76 & 64.86 \\
zom\_Latn & 50.45 & 42.34 & 44.14 & 43.24 & 48.65 \\
zsm\_Latn & 58.56 & 59.46 & 63.96 & 55.86 & 66.67 \\
zul\_Latn & 46.85 & 42.34 & 46.85 & 46.85 & 51.35 \\
\midrule
all & 44.07 & 40.98 & 43.98 & 43.24 & 48.89 \\
\bottomrule
\end{tabular}
}
\caption{Detailed results on \taxi (Part VI). 3-shot results are presented.}
\label{taxi6}
\end{table*}

\end{document}